\documentclass[runningheads, orivec]{llncs}
\usepackage[T1]{fontenc}
\usepackage{graphicx}
\usepackage{amsmath}
\usepackage{amssymb}
\usepackage{tabularx}
\usepackage{subcaption}
\usepackage[hidelinks]{hyperref}
\usepackage{color}

\begin{document}
\title{Decentralized Evolution of Hexapod Gaits with Independent Leg Controllers}
\titlerunning{Decentralized Evolution of Hexapod Gaits with Independent Leg Controllers}
%
\author{Gary B. Parker\inst{1}\orcidID{0009-0001-3870-1190} \and
John Asaro\inst{1}\orcidID{0009-0001-0723-4921} \and
Jim O'Connor\inst{1}\orcidID{0009-0008-9917-5682}}
\authorrunning{Parker et al.}
%
\institute{{Connecticut College, New London CT 06320-4125, USA}
\\
\email{\{parker, jasaro, joconno2\}@conncoll.edu}}
\maketitle              
\begin{abstract}

This paper presents a novel approach to hexapod locomotion by evolving each leg's gait independently through a decentralized evolutionary algorithm. Using the Webots simulator and the Mantis hexapod robot, we optimize individual leg controllers without centralized coordination, allowing emergent behaviors to drive the development of efficient, coordinated locomotion. Our decentralized method is benchmarked against cooperative coevolution, demonstrating improved efficacy in generating stable and adaptive gaits while showing interesting emergent coordination. By enabling independent evolution of leg controllers, this method reduces the complexity of gait optimization and highlights the potential of decentralized strategies for scalable and adaptive robotic systems.

\end{abstract}

\section{Introduction}
\label{sec:introduction}
Hexapod robots, with their inherently stable and redundant locomotion capabilities, have long been a focal point of research in robotics. Their ability to traverse uneven and unpredictable terrain makes them suitable for a range of applications, from planetary exploration to disaster response \cite{Disaster}. However, developing efficient and robust gaits for hexapods presents a significant challenge due to the high-dimensional nature of their control spaces and the need for precise coordination across multiple legs. This problem has traditionally been approached using centralized control mechanisms  \cite{byrd_devries_1990}  or evolutionary algorithms designed to optimize the entire gait as a unified entity \cite{parker1996cyclic}. While such methods have produced effective solutions, they can suffer from scalability issues and are sensitive to disruptions in coordination. Decentralized approaches, in which the control of each leg is evolved independently, represent a promising alternative that could simplify the optimization process while leveraging emergent behavior to achieve coordinated locomotion \cite{Angliss2023CoevolvingHL}.

Early attempts at hexapod gait generation were largely inspired by biological observations, relying on manually designed gaits that mimicked the patterns observed in insects and other multi-legged animals \cite{hexa-stickbug}. These gaits were typically pre-programmed, based on fixed trajectories or predefined phase offsets between legs \cite{Song1984KinematicOD}. While effective for simple tasks and controlled environments, such methods lacked adaptability, making them unsuitable for complex, real-world scenarios where robots must dynamically adjust to environmental changes or unexpected perturbations.

The use of evolutionary algorithms (EAs) marked a noticeable shift in gait optimization strategies. Centralized EAs treated the entire gait as a single monolithic entity, evolving parameters such as joint angles, phase offsets, and amplitudes to maximize locomotion efficiency \cite{cpgs-ga}. These approaches demonstrated the potential of hexapod gait generation, particularly in producing novel and effective locomotion strategies. However, the high-dimensional search spaces associated with centralized generation brought forth challenges in scaling, particularly when applied to robots with more complex morphologies \cite{Modular-Robots}.

In response, cooperative coevolution emerged as an alternative framework that decomposed the gait optimization problem into smaller, more manageable subproblems \cite{coop-coevolution-1994}. Under this paradigm, individual components of the gait were treated as separate entities that evolved in tandem. By focusing on the interactions between these components, cooperative coevolution achieved a balance between modularity and interdependence, allowing for the emergence of coordinated gaits while reducing the computational burden of high-dimensional optimization. Despite its success, cooperative coevolution relies heavily on the assumption that interactions between components can be effectively managed during optimization. This introduces additional complexity, particularly in ensuring that components evolve in a manner that promotes overall system coherence.

The concept of decentralized control has its roots in distributed systems and swarm robotics, where individual agents operate independently, often with minimal or no communication between them. Decentralized evolutionary algorithms leverage this principle by allowing each component of a system to evolve independently, guided only by local objectives or environmental feedback. Such approaches have been applied successfully in various domains, including swarm behavior optimization \cite{swarm}, distributed sensor networks \cite{sensor-networks}, and modular robotics \cite{saldana2017decentralized}. In these contexts, decentralized evolution has demonstrated advantages such as robustness to individual component failures, scalability to complex systems, and the ability to exploit emergent behavior to achieve global objectives.

This paper introduces a novel decentralized evolutionary framework for hexapod gait generation, wherein the controller for each leg evolves independently without knowledge of the other legs. By relying solely on emergent behavior to achieve coordination, this approach departs from traditional methods that impose explicit inter-leg communication or centralized control. Each leg’s controller is optimized using a genetic algorithm, with fitness determined by the robot’s overall locomotion performance. This setup not only simplifies the optimization process but also provides a natural mechanism for robustness and adaptability, as the independent controllers can autonomously adapt to changes or failures in other legs.

The proposed approach is evaluated in a simulated environment using the Webots simulator \cite{michel2004cyberbotics} and the Mantis hexapod robot. To benchmark its performance, we compare our decentralized evolution algorithm against the previously mentioned cooperative coevolution method. Our results demonstrate that decentralized evolution significantly outperforms this method in producing efficient and coordinated gaits. Moreover, the decentralized framework offers distinct advantages in terms of modularity and scalability for legged robotic locomotion.

\section{Simulation Environment}
\label{sec:env}
The experiments were conducted using the Webots simulator, a widely used platform for simulating robotic systems in dynamic and physics-based environments. Webots provides a realistic simulation of physical interactions, including friction, inertia, and collision dynamics, making it particularly suitable for the study of legged robotic locomotion. Key environmental parameters were carefully calibrated to balance realism and computational feasibility. The simulation was conducted under a standard gravity and physics model, with the robot's interactions constrained by friction coefficients suitable for moderate grip surfaces. These constraints ensure that locomotion efficiency and stability were primarily functions of the evolved gaits, rather than products of environmental attributes. 

\begin{figure}
    \centering
    \includegraphics[width=0.75\linewidth]{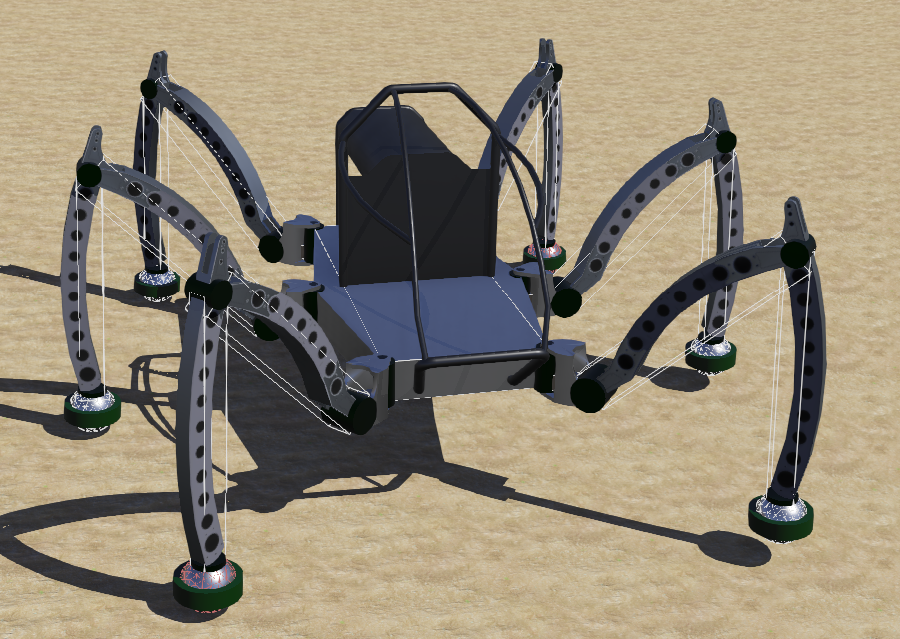}
    \caption{The Mantis hexapod robot within the Webots simulation environment. The model features six legs with three degrees of freedom each.}
    \label{fig:enter-label}
\end{figure}
\subsection{The Mantis Hexapod}
\label{sec:mantis}
The Mantis hexapod robot, originally developed by Matt Denton of Micromagic Systems \cite{denton_2016_mantis}, was chosen as the foundational model for this study due to its biologically inspired design and it's inclusion in Webots as a standard model. As the largest operational hexapod ever constructed, the physical Mantis robot is powered by a Perkins 2.2L turbo diesel engine and employs hydraulic actuators to drive its 18 degrees of freedom (DOF). However, for this study, a simulated version of the Mantis was utilized within the Webots simulator to evaluate the performance of decentralized evolutionary algorithms for gait generation under realistic conditions.

The simulated Mantis closely mirrors the structure and functionality of its physical counterpart. Each of the six legs features three degrees of freedom: the coxa joint facilitates horizontal movement, the femur joint controls vertical positioning, and the tibia joint determines ground interaction. These 18 joints, fundamental to hexapod locomotion, were modeled with biologically plausible constraints to ensure realistic gait evolution. Specifically, the coxa joint range was set to ±90 degrees, the femur joint to +60 to -75 degrees, and the tibia joint to -120 to -20 degrees. This reflects a reasonable estimate of the mechanical capabilities of the original Mantis robot, while also not constraining the robot so much that a stable and efficient gait is guaranteed. We break each 32 millisecond interval of simulation time into a single \emph{timestep}. The robot is able to manipulate all of its motors at a given timestep, but any motor can only move ±5 degrees from its position at the previous timestep.

The simulated mantis was equipped with a GPS placed directly in the center of mass of the robot body. This allowed us to track the robots position in real time, which factored into the robot fitness calculations.

Each leg in the simulation operates independently, governed by its own controller optimized through the decentralized evolutionary framework described in subsequent sections. By using the standard Mantis hexapod model, we can ensure that the results obtained from the simulation are both experimentally consistent and potentially applicable to physical robotic systems. 

\section{Decentralized Evolutionary Framework}
Our decentralized evolutionary framework introduces a novel approach to hexapod gait generation, leveraging the independent evolution of controllers for each leg without any centralized coordination mechanism. By treating each leg as an autonomous entity, this framework exploits the emergent behaviors arising from local optimizations to achieve globally coordinated locomotion. The framework consists of two primary components: Central Pattern Generators (CPGs) as the underlying control model for each leg and a genetic algorithm (GA) for optimizing the parameters of the CPGs.

\subsection{Central Pattern Generators}

Central Pattern Generators are biologically inspired neural circuits that produce rhythmic outputs without requiring rhythmic inputs. These neural mechanisms, found in a variety of biological organisms, are integral to generating rhythmic behaviors such as walking, swimming, and breathing. The fundamental property of CPGs is their ability to produce stable and adaptive oscillatory signals with minimal external feedback, which has made them a subject of extensive research in both neuroscience and robotics \cite{marder2001central}.

The study of CPGs in robotics can be traced back to their biological origins, where they were first identified as the neural mechanisms controlling rhythmic movements in animals. Early computational models of CPGs were inspired by observations of simple organisms such as lampreys and leeches, whose undulating swimming motions are governed by networks of coupled oscillators in their central nervous systems \cite{eisenhart2000central}. These initial studies demonstrated that complex rhythmic behaviors could emerge from relatively simple neural circuits.

In robotics, CPGs are useful as a control strategy due to their intrinsic properties of stability, robustness, and adaptability. Unlike traditional kinematic approaches that rely on explicit trajectory planning, CPGs offer an easily decentralized and distributed solution, reducing the computational burden and improving fault tolerance. Early implementations of CPGs in robotic systems focused on swimming robots and quadrupedal locomotion, where coupled oscillator networks were used to produce coordinated rhythmic motions \cite{reeve1994control}. These systems demonstrated the potential of CPG-based control to handle uneven terrain, adapt to perturbations, and recover from partial failures, making them particularly well-suited for legged robots.

The CPG model in this study is parameterized to generate sinusoidal joint trajectories defined by amplitude, phase, and offset values. These parameters are evolved for each joint of the hexapod robot to produce effective locomotion patterns. The motion of each joint is given by:
\begin{equation}
\theta_i(t) = A_i \sin(2\pi f t + \phi_i) + O_i,
\label{eq:joint_position}
\end{equation}
where \(\theta_i(t)\) represents the angular position of joint \(i\) at time \(t\), \(A_i\) is the amplitude of oscillation for joint \(i\), \(f\) is the fixed frequency of oscillation shared across all joints, \(\phi_i\) is the phase offset for joint \(i\), and \(O_i\) is the positional offset for joint \(i\).

These parameters are optimized independently for each of the hexapod’s legs, adhering to the decentralized nature of the framework. This approach ensures that each leg evolves autonomously, with emergent coordination arising from the shared fitness evaluation.

\subsection{Genetic Algorithm for Parameter Optimization}
To optimize the parameters of the CPGs for each leg, we employ a genetic algorithm (GA) tailored to the decentralized framework. Each leg’s controller is treated as an individual with its own population, and the GA evolves the parameters of the sinusoidal trajectories independently. The parameters optimized by the GA include the amplitude \(A_i\), which determines the range of motion for the coxa, femur, and tibia joints; the phase \(\phi_i\), which controls the timing of oscillations relative to other joints; and the offset \(O_i\), which defines the baseline position of each joint.
The fitness of each leg’s controller is evaluated based on its contribution to the robot’s overall locomotion performance. The fitness function is defined as:
\begin{equation}
F = d
\end{equation}
where \(F\) is the fitness score and \(d\) represents the maximum distance traveled by the hexapod on the X-axis from its starting position. This is determined by using the robots GPS, sampling the robots position on the X-axis at every timestep, and calculating the euclidean distance from the starting position.

The GA employs stochastic proportionate selection to choose parent individuals for reproduction based on their relative fitness. Reproduction is facilitated through uniform crossover, which combines the parameters of two parent individuals to explore novel parameter combinations. Single point mutation is used to promote diversity within the population and prevent premature convergence.
The evolutionary process begins with the initialization of each population using random parameter vectors. Each individual’s fitness is then computed, and individuals are selected for reproduction based on their fitness. Crossover and mutation operators are applied to generate a new population, and the process is repeated for a predefined and empirically validated number of generations to ensure convergence. By optimizing the CPG parameters through this GA-based framework, the controller of each leg evolves to maximize locomotion efficiency and stability. This decentralized approach enables emergent coordination between legs, resulting in globally effective gaits without requiring centralized control.

\section{Testing Methodology}
\label{sec:methodology}
To evaluate the performance of the decentralized evolutionary framework, we conducted a comparative analysis against a cooperative coevolution baseline. Both approaches utilize a population size of 300 individuals per generation, with 50 individuals allocated per leg. Each individual encodes a set of sinusoidal control parameters for one leg using a chromosome consisting of three genes, one per joint, with each gene represented by three real-valued parameters.

In the decentralized framework, each generation comprises 50 gaits formed by sampling one individual from each leg’s population. Each of these 50 complete gaits is then evaluated in simulation, and the resulting locomotion performance is used as the fitness signal for all participating leg controllers. In contrast, the cooperative coevolution framework evaluates each leg controller individually by combining it with the highest-performing individuals from the previous generation for the remaining legs. This nested evaluation process results in a sixfold increase in simulation runs per generation, leading to significantly longer training times.

Fitness is computed by measuring the furthest displacement along the X-axis achieved by the hexapod during a 20-second evaluation window. The position of the robot is recorded at each timestep using its onboard GPS module, and the maximum Euclidean displacement from the initial position is retained as the final fitness value.

To ensure fair comparison, all evolutionary algorithms were allowed to run for an equivalent number of simulation steps. The decentralized and centralized baselines were run for 3000 generations, while cooperative coevolution was executed for 500 generations. Since each generation of cooperative coevolution involves six times more evaluations, we define an \emph{effective generation} as a single pass over 300 complete gaits, equivalent to one generation in the decentralized model or six in the cooperative coevolution model. Performance metrics are reported in terms of effective generations to normalize for evaluation cost.

To test robustness under partial actuator failure, we conducted ablation experiments where one leg was disabled throughout training. Each ablation scenario was run for 3000 effective generations in the decentralized model and 600 effective generations in the cooperative coevolution model to maintain parity in total evaluations.

\begin{figure}[t!]
    \centering
    \includegraphics[width=1\linewidth]{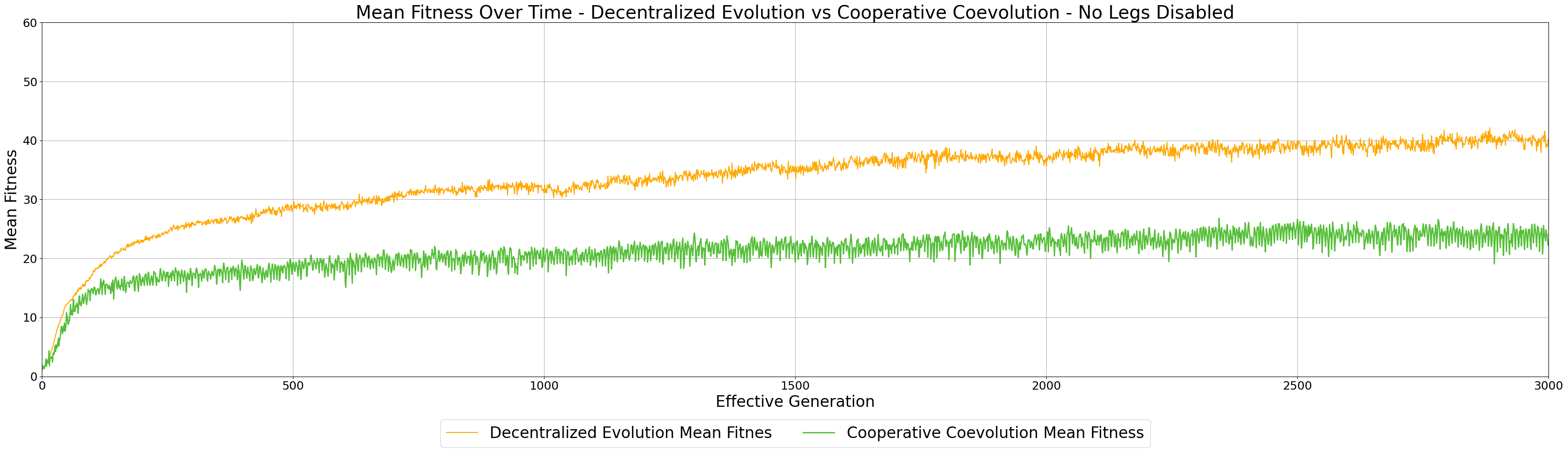}
    \caption{Mean Fitness Over Time (No Legs Disabled): Average fitness of the population across 3000 effective generations with all six legs operational.}
    \label{fig:Mean Fitness Overtime - No Legs Disabled}
\end{figure}
\begin{figure}[t!]
    \centering
    \includegraphics[width=1\linewidth]{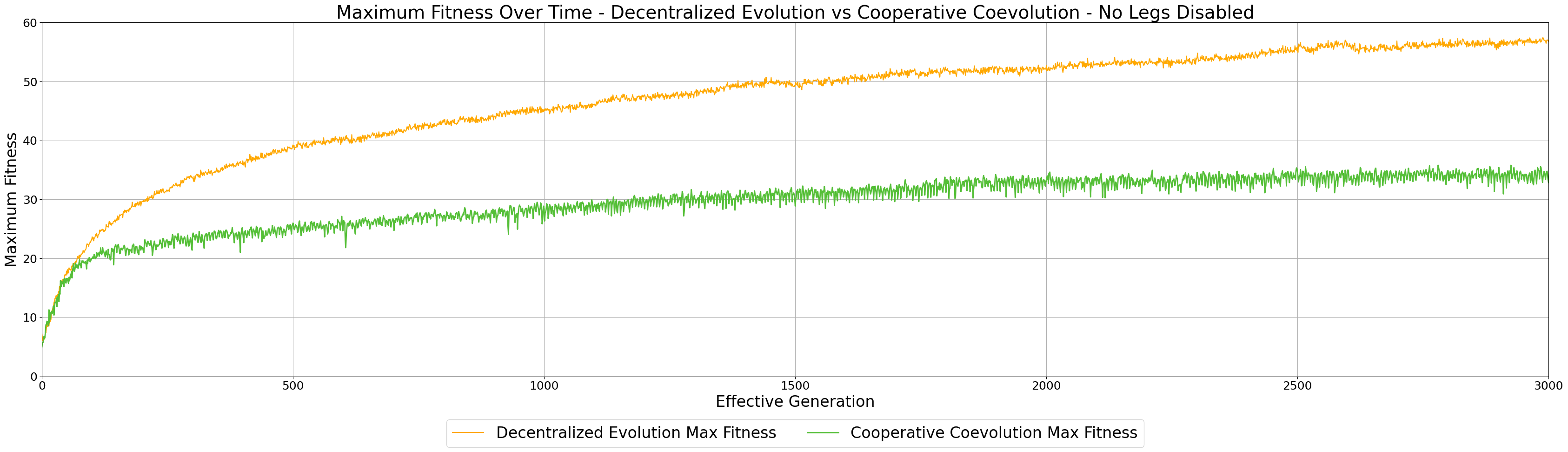}
    \caption{Maximum Fitness Over Time (No Legs Disabled): Best individual fitness observed in each generation across 10 independent runs with no leg ablation.}
    \label{fig:Maximum Fitness Overtime - No Legs Disabled}
\end{figure}

\section{Results}
\label{results}

Performance metrics were averaged over 10 independent runs for each experimental condition. Figures \ref{fig:Mean Fitness Overtime - No Legs Disabled} and \ref{fig:Maximum Fitness Overtime - No Legs Disabled} show the mean and maximum fitness values, respectively, over the course of evolution when all six legs were operational. In both cases, the decentralized framework consistently outperforms cooperative coevolution in terms of convergence speed and final fitness achieved.

To assess robustness to partial actuator failure, we conducted a series of leg ablation experiments, each disabling a single leg during the entirety of training. The results of these experiments are presented in the Appendix (Figures \ref{fig:appendix_leg0_mean} through \ref{fig:appendix_leg4_max}). Across all ablation scenarios, the decentralized method consistently achieves higher locomotion fitness and demonstrates greater adaptability than the cooperative coevolution baseline. These findings reinforce the conclusion that decentralized evolution produces controllers that are not only efficient but also robust to structural impairments.

In all experiments, decentralized evolution exhibits smoother convergence and reduced performance variance compared to cooperative coevolution. These findings support the hypothesis that emergent coordination can arise from fully decentralized optimization and that such coordination is inherently more resilient to disruptions in the control structure.

\section{Conclusion}
This work introduces and evaluates a decentralized evolutionary framework for hexapod gait generation, where each leg is controlled by an independently evolved CPG-based controller. Comparative experiments against cooperative coevolution demonstrate that the decentralized approach achieves higher locomotion pemy. Unfortunately this method can be too
computationally intense to be carried out on board the robot. A
system of learning that can be carried out offline and then
downloaded to the onboard contrrformance with faster convergence while also exhibiting superior robustness to leg ablation.

Although we account for evaluation cost by reporting results in terms of effective generations, it is worth noting that decentralized evolution continues to outperform cooperative coevolution even in early stages of training, suggesting efficiency gains beyond simple run-time equivalence. These findings highlight the potential of decentralized methods for scalable, resilient control in multi-legged robotic systems.

Future work will investigate the performance of this framework on physical hexapod platforms and explore extensions to hybrid models that combine decentralized evolution with limited local communication. Additional comparisons against fully centralized monolithic controllers would further contextualize the benefits of this approach within the broader landscape of robot gait optimization.

\bibliographystyle{splncs04}
\bibliography{hexapod_citations}

\vspace{100px}
\appendix
\section*{Appendix: Leg Ablation Results}

\begin{figure}[ht]
    \centering
    \includegraphics[width=\linewidth]{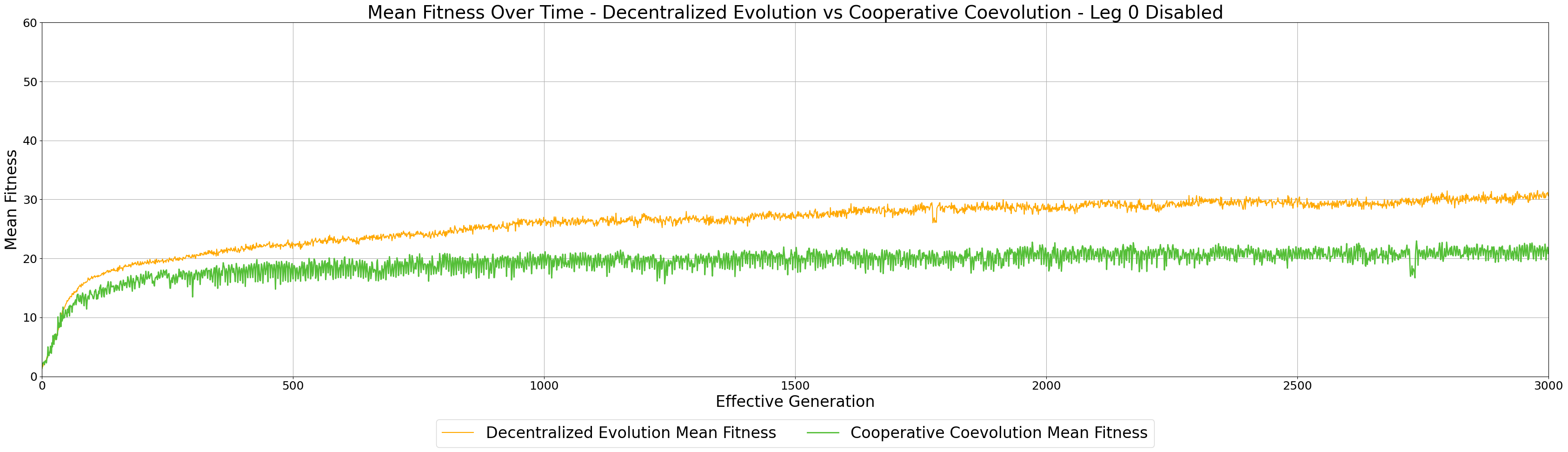}
    \caption{Mean fitness over time with leg 0 disabled across 10 independent runs.}
    \label{fig:appendix_leg0_mean}
\end{figure}

\begin{figure}[ht]
    \centering
    \includegraphics[width=\linewidth]{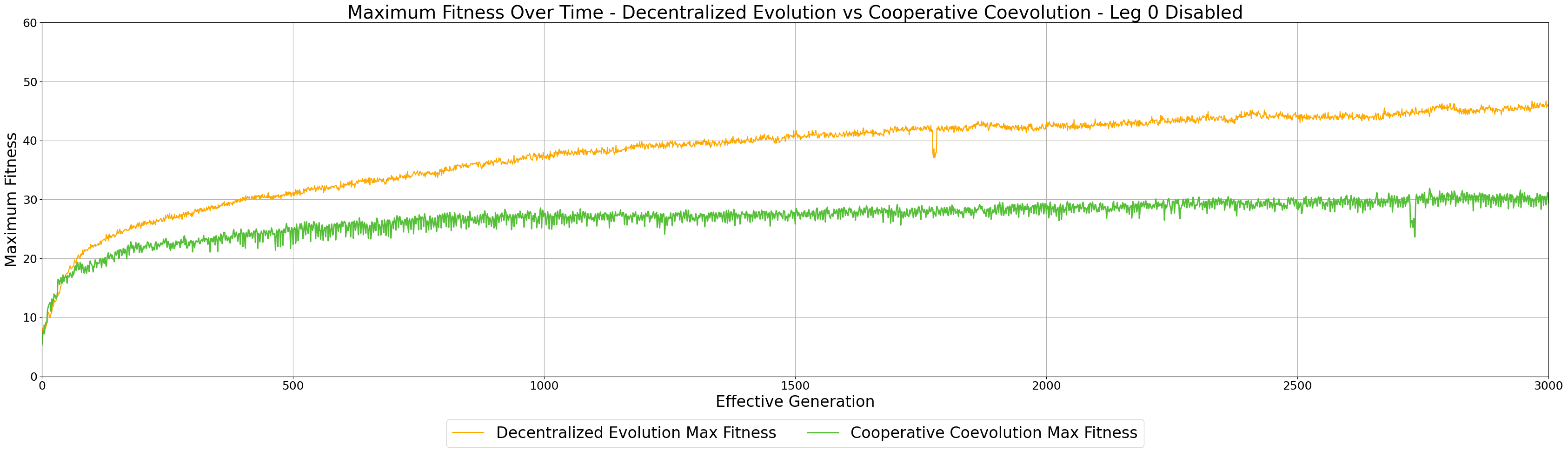}
    \caption{Maximum fitness over time with leg 0 disabled across 10 independent runs.}
    \label{fig:appendix_leg0_max}
\end{figure}

\begin{figure}[ht]
    \centering
    \includegraphics[width=\linewidth]{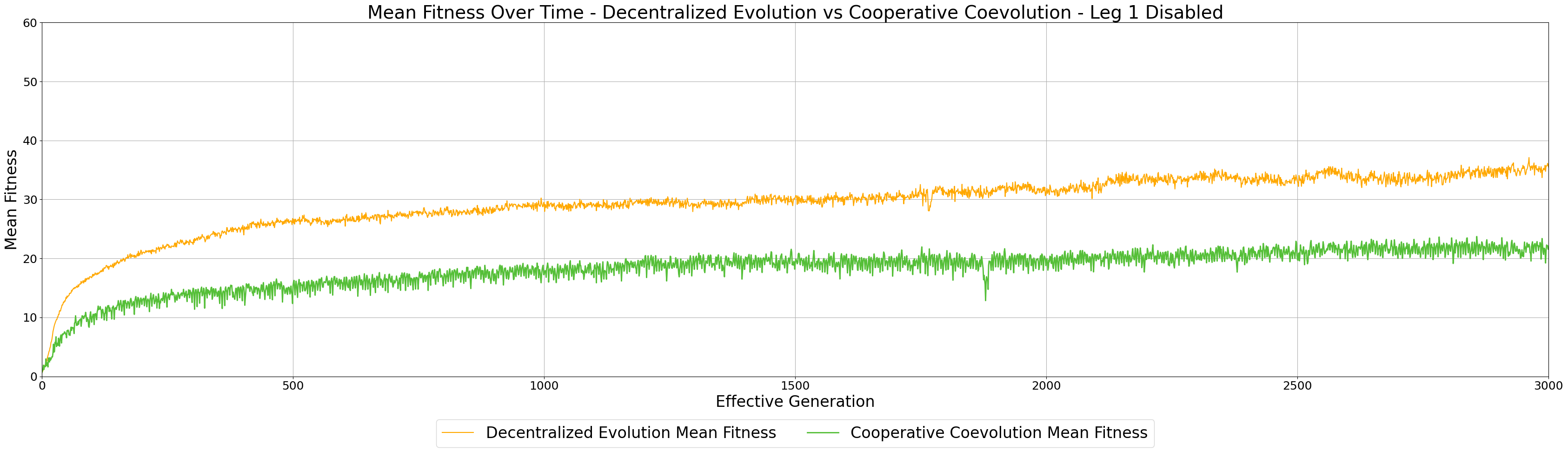}
    \caption{Mean fitness over time with leg 1 disabled across 10 independent runs.}
    \label{fig:appendix_leg1_mean}
\end{figure}

\begin{figure}[ht]
    \centering
    \includegraphics[width=\linewidth]{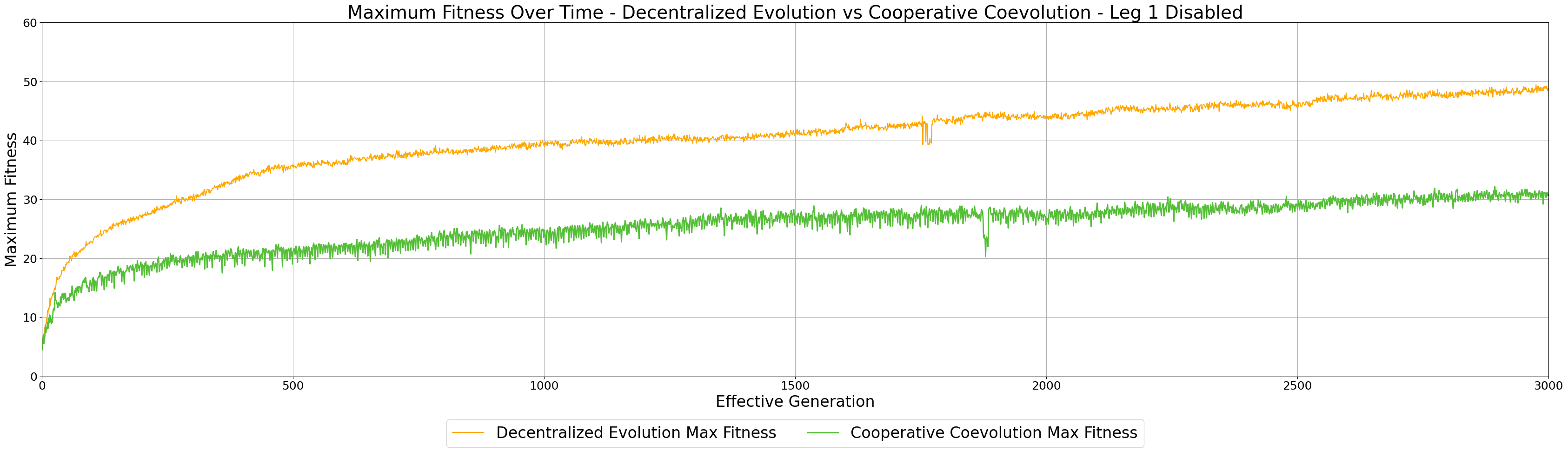}
    \caption{Maximum fitness over time with leg 1 disabled across 10 independent runs.}
    \label{fig:appendix_leg1_max}
\end{figure}

\begin{figure}[ht]
    \centering
    \includegraphics[width=\linewidth]{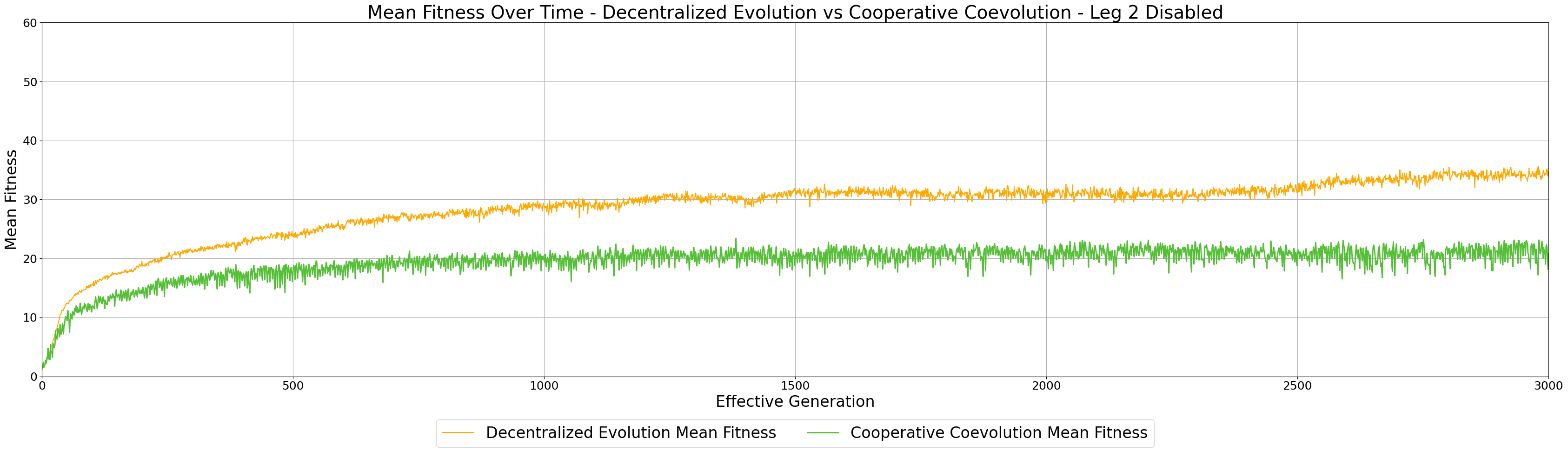}
    \caption{Mean fitness over time with leg 2 disabled across 10 independent runs.}
    \label{fig:appendix_leg2_mean}
\end{figure}

\begin{figure}[ht]
    \centering
    \includegraphics[width=\linewidth]{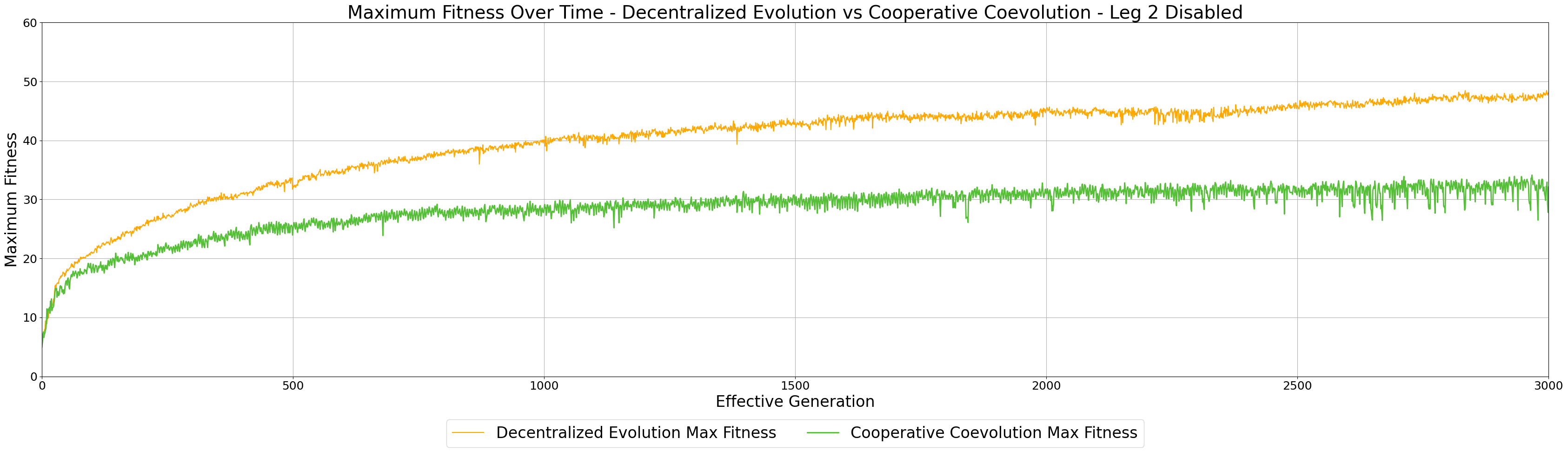}
    \caption{Maximum fitness over time with leg 2 disabled across 10 independent runs.}
    \label{fig:appendix_leg2_max}
\end{figure}

\begin{figure}[ht]
    \centering
    \includegraphics[width=\linewidth]{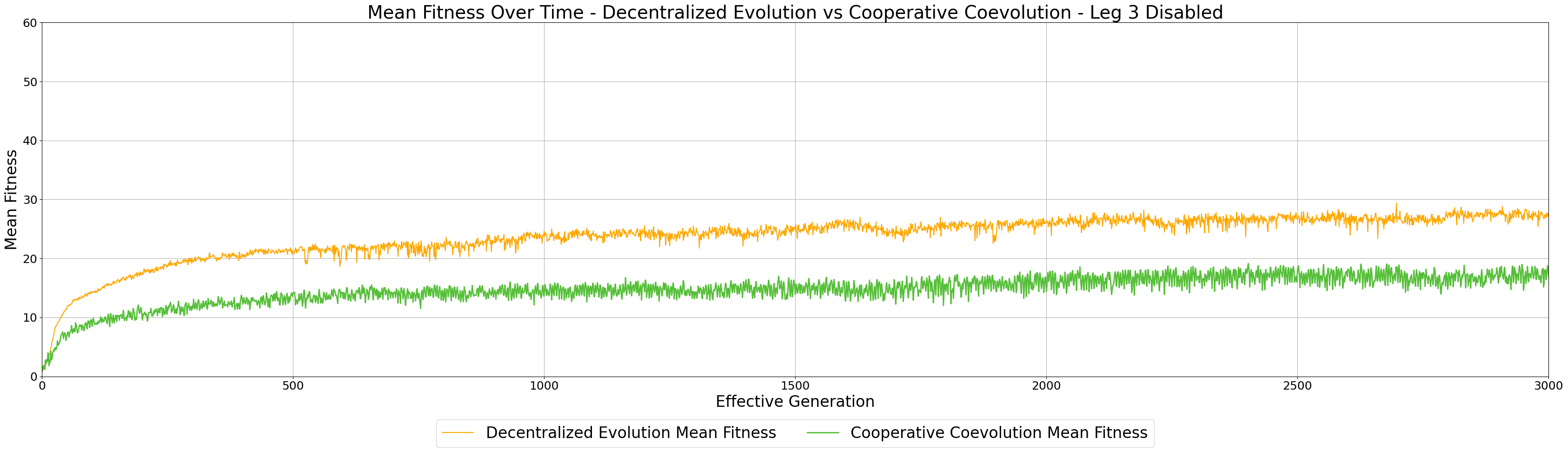}
    \caption{Mean fitness over time with leg 3 disabled across 10 independent runs.}
    \label{fig:appendix_leg3_mean}
\end{figure}

\begin{figure}[ht]
    \centering
    \includegraphics[width=\linewidth]{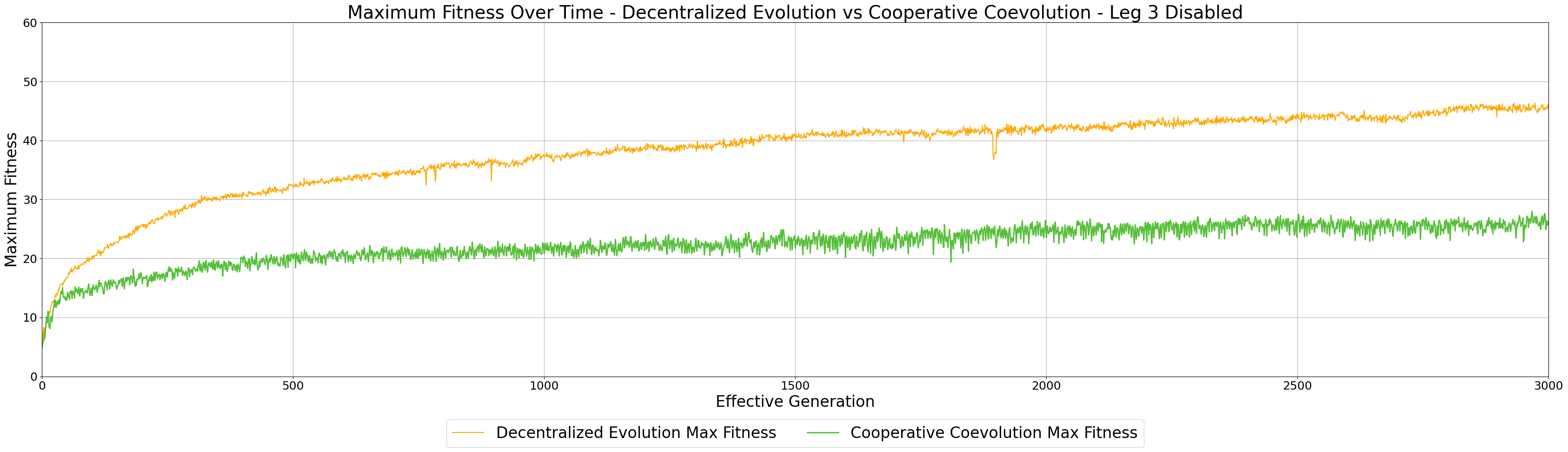}
    \caption{Maximum fitness over time with leg 3 disabled across 10 independent runs.}
    \label{fig:appendix_leg3_max}
\end{figure}

\begin{figure}[ht]
    \centering
    \includegraphics[width=\linewidth]{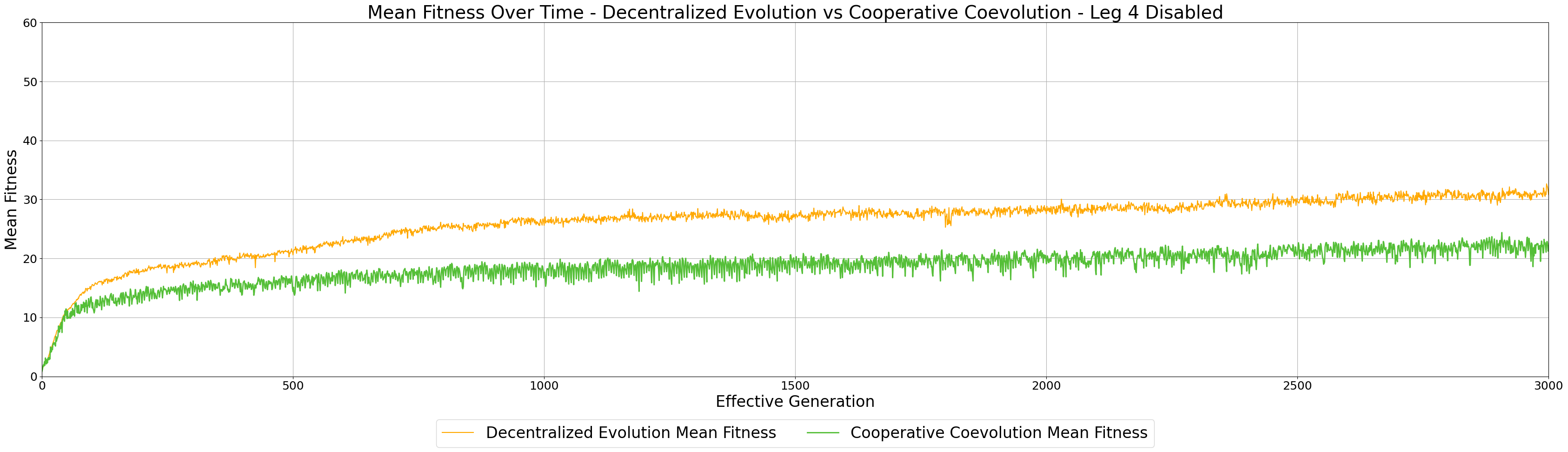}
    \caption{Mean fitness over time with leg 4 disabled across 10 independent runs.}
    \label{fig:appendix_leg4_mean}
\end{figure}

\begin{figure}[ht]
    \centering
    \includegraphics[width=\linewidth]{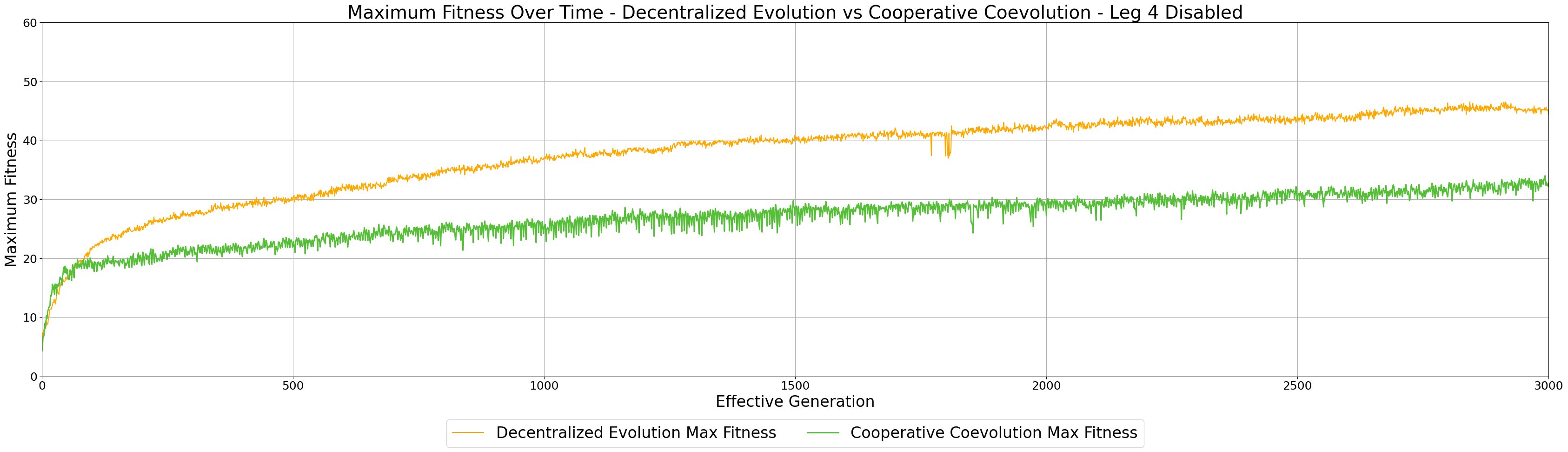}
    \caption{Maximum fitness over time with leg 4 disabled across 10 independent runs.}
    \label{fig:appendix_leg4_max}
\end{figure}

\begin{figure}[ht]
    \centering
    \includegraphics[width=\linewidth]{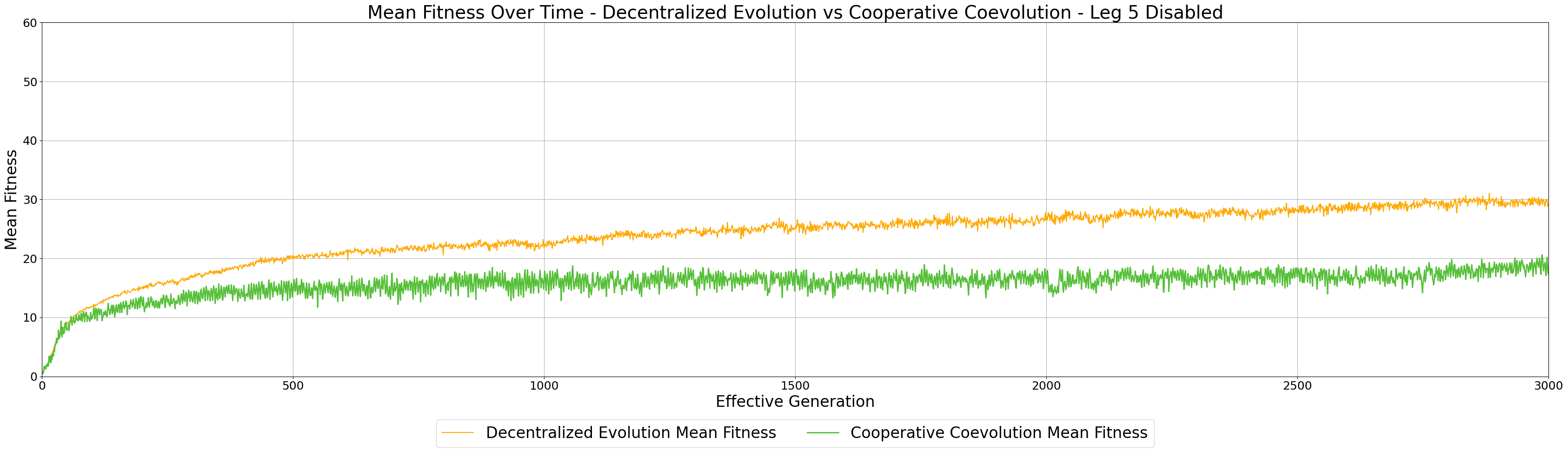}
    \caption{Mean fitness over time with leg 5 disabled across 10 independent runs.}
    \label{fig:appendix_leg4_mean}
\end{figure}

\begin{figure}[ht]
    \centering
    \includegraphics[width=\linewidth]{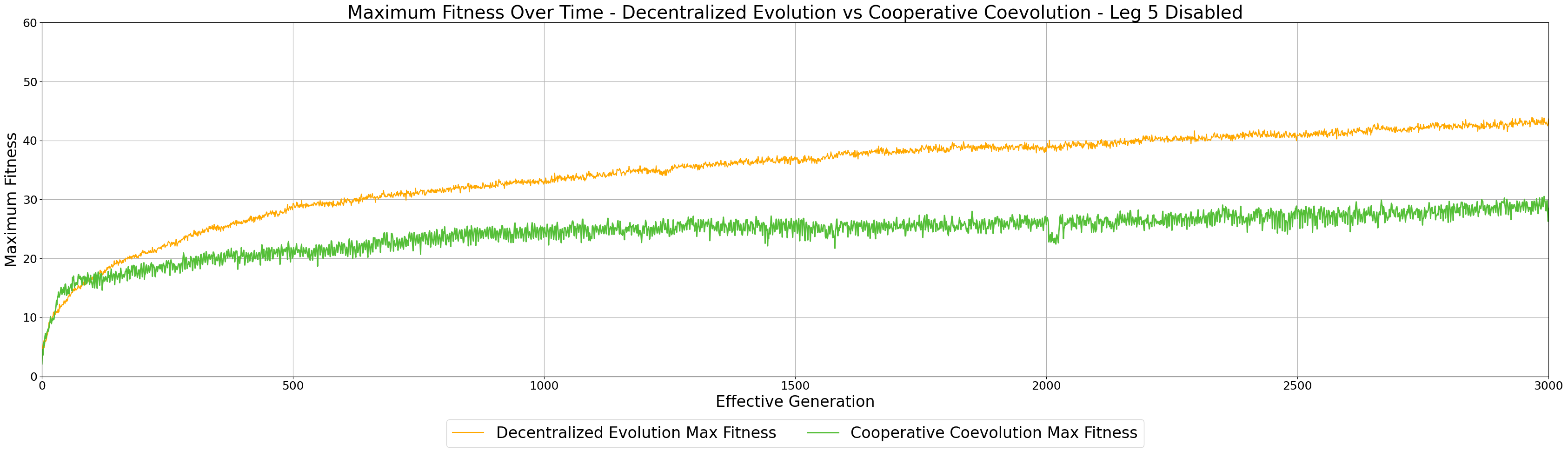}
    \caption{Maximum fitness over time with leg 5 disabled across 10 independent runs.}
    \label{fig:appendix_leg4_max}
\end{figure}

\end{document}